\documentclass[runningheads]{llncs}
\usepackage{multicol}
\usepackage{multirow}
 
\usepackage{eccv}

\def\figvspacemid{\vspace{-10pt}}
\def\tablevspacemid{\vspace{-6pt}}\def\figvspacebottom{\vspace{-0pt}}

\usepackage{eccvabbrv}

\usepackage{graphicx}
\usepackage{booktabs}

\usepackage[accsupp]{axessibility}  

\usepackage[pagebackref,breaklinks,colorlinks]{hyperref}

\usepackage{orcidlink}

\usepackage{tabularx}
\newcount\K
\def\bla#1{
\K=0 \loop\ifnum\K<#1
{\textcolor[gray]{0.9}{{\it bla bla bla bla bla bla bla bla bla bla bla bla bla bla bla}}}
\advance\K by1\repeat
}

\begin{document}

\title{Rethinking Image Super-Resolution from Training Data Perspectives} 

\titlerunning{Rethinking Image Super-Resolution from Training Data Perspectives}

\author{Go Ohtani\inst{1,2}\orcidlink{0009-0009-6289-6272} \and
Ryu Tadokoro\inst{1}\orcidlink{0009-0001-9473-3832}\and
Ryosuke Yamada\inst{1,3}\orcidlink{0000-0002-2154-8230}\and
Yuki M. Asano\inst{4}\orcidlink{0000-0002-8533-4020}\and\\
Iro Laina\inst{5}\orcidlink{0000-0001-8857-7709}\and
Christian Rupprecht\inst{5}\orcidlink{0000-0003-3994-8045}\and
Nakamasa Inoue\inst{6,1}\orcidlink{0000-0002-9761-4142}\and
Rio Yokota\inst{6,1}\orcidlink{0000-0001-7573-7873}\and
Hirokatsu Kataoka\inst{1}\orcidlink{0000-0001-8844-165X}\and
Yoshimitsu Aoki\inst{2}\orcidlink{0000-0001-7361-0027}}

\authorrunning{G. Ohtani et al.}

\institute{
National Institute of Advanced Industrial Science and Technology (AIST)\and
Keio University \and
University of Tsukuba\and
University of Amsterdam\and 
University of Oxford\and
Tokyo Institute of Technology
}

\maketitle

\begin{abstract}
In this work, we investigate the understudied effect of the training data used for image super-resolution (SR). Most commonly, novel SR methods are developed and benchmarked on common training datasets such as DIV2K and DF2K. However, we investigate and rethink the training data from the perspectives of diversity and quality, {thereby addressing the question of ``How important is SR training for SR models?''}. To this end, we propose an automated image evaluation pipeline. With this, we stratify existing high-resolution image datasets and larger-scale image datasets such as ImageNet and PASS to compare their performances. We find that datasets with (i) low compression artifacts, (ii) high within-image diversity as judged by the number of different objects, and (iii) a large number of images from ImageNet or PASS all positively affect SR performance. We hope that the proposed simple-yet-effective dataset curation pipeline will inform the construction of SR datasets in the future and yield overall better models. Code is available at: \href{https://github.com/gohtanii/DiverSeg-dataset}{https://github.com/gohtanii/DiverSeg-dataset}


\end{abstract}


\section{Introduction}
\label{sec:intro}
Image super-resolution (SR) aims to reconstruct high-resolution images from low-resolution images.
It has been considered as one of the most fundamental tasks in the field of computer vision, with applications ranging from autonomous driving to medical imaging.
Deep learning methods have led to significant advances in SR over the last decade, focusing primarily on improvements in the neural network architectures.
Early SR models rely on convolutional neural networks (CNNs)~\cite{dong2014learning, kim2016accurate, dong2016accelerating, shi2016real, ledig2017photo, tong2017image, lim2017enhanced, wang2018esrgan, zhang2018residual, zhang2018image, zhang2019residual}. Recent innovations have given rise to transformer-based SR models~\cite{chen2021pre, li2021efficient, liang2021swinir, zhang2022swinfir, zhang2022efficient, chen2023activating, Zhou_2023_ICCV}, which have consistently improved the performances.

\begin{figure*}[t]
\centering
\includegraphics[width=\linewidth]{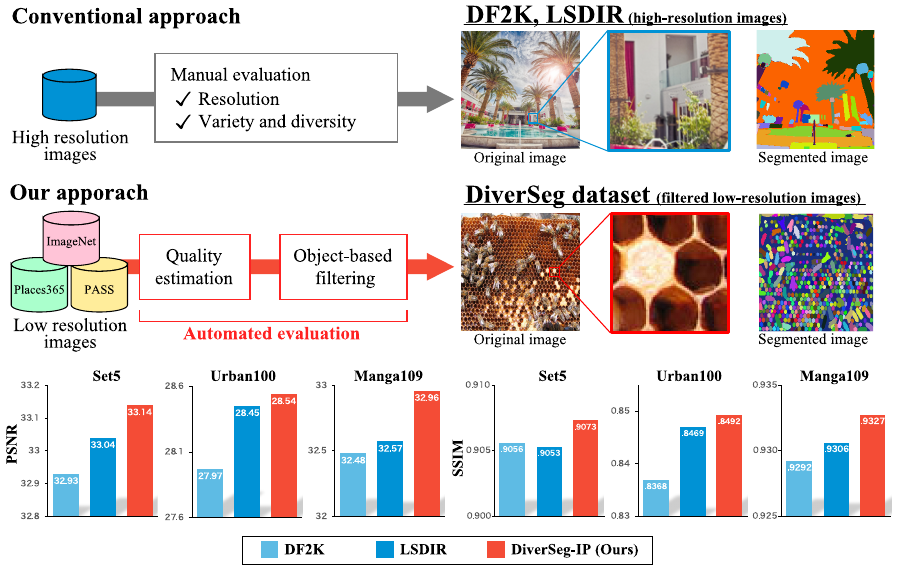}
\figvspacemid
\caption{
We propose an automated image evaluation pipeline to curate a dataset for training SR models.
The obtained dataset, namely DiverSeg, consists of low-resolution but high-quality images with many object regions.
SR models trained on DiverSeg outperform those trained on high-resolution image datasets such as DF2K and LSDIR.}
\label{fig:teaser}
\figvspacebottom
\end{figure*}

With the improvement of neural network architectures, the importance of training datasets has also increased, as discussed in~\cite{Li_2023_CVPR}. 
Examples of high-resolution datasets include DIV2K~\cite{agustsson2017ntire} and Flickr2K~\cite{timofte2017ntire}.
The combined dataset of these two, referred to as DF2K, is often utilized for training SR models.
Most recently, LSDIR~\cite{Li_2023_CVPR} has been proposed, which consists of 84,991 high-resolution images.
It has been confirmed that training on large high-resolution datasets contributes significantly to performance improvement~\cite{wang2018esrgan, liang2021swinir, Zhou_2023_ICCV, li2023ntire, zhang2023ntire} 

The conventional approach to constructing datasets relies on a manual evaluation step, where the following two perspectives are most commonly considered:
\begin{enumerate}
\item \textbf{Resolution and quality~\cite{Li_2023_CVPR,yang2023hq}.} This perspective focuses on the pixel density of images. Images that do not meet the specified resolution threshold are excluded. Typically, HD, 2K, and 4K images are used to create a dataset. After the initial automatic filtering based on image size, the details of each image are manually evaluated to identify and exclude compressed images.

\item \textbf{Variety and diversity~\cite{li2023feature}.} This includes diversity in subjects ({\it e.g.}, people, landscapes, urban scenes), lighting conditions, colors, textures, and other photographic elements. A diverse dataset is said to help train a model that is robust and performs well across a wide range of domains.
\end{enumerate}
Datasets constructed from these perspectives have been shown to significantly improve the performance of SR models. However, they also pose challenges in scaling the datasets, as collecting uncompressed high-resolution images is difficult and costly.
\setcounter{footnote}{0}

To address this limitation, this paper rethinks these perspectives and proposes {Diverse Segmentation dataset (DiverSeg)} , a low-resolution{\footnote{We define images with a resolution lower than HD as low-resolution images.}} yet effective image dataset for training SR models.
As shown in Figure~\ref{fig:teaser}, the dataset is constructed by applying filtering to a large set of low-resolution images, such as ImageNet-1k~\cite{deng2009imagenet} and PASS~\cite{asano21pass}.
In experiments, we demonstrate that models trained on DiverSeg outperform those trained on high-resolution image datasets such as DF2K and LSDIR. Based on this finding, our contributions are summarized as follows.

\noindent \textbf{1) Rethinking the resolution perspective.}
High-resolution images have been considered to be necessary for training SR models. In this work, we challenge this traditional perspective and show that SR models can be trained without high-resolution images.
Specifically, we introduce a method to estimate image quality based on the kernel density estimation over blockiness values~\cite{bhardwaj2018jpeg} that estimates the quantity distribution of blocking artifacts. {We demonstrate that low-resolution images with high quality, indicated by reduced artifacts, can improve the performance of SR models. We also thoroughly analyze the impact of image quality on SR performance.}

\noindent \textbf{2) Rethinking the diversity perspective.}
When constructing datasets for training  SR models, images containing only a small number of objects are often implicitly excluded during the manual resolution evaluation process. This is because evaluators typically focus on the details of objects or small objects in the images.
{Therefore, we explicitly calculate the number of objects in images and analyze how this number affects SR performance.}
In {our} experiments, we show that constructing datasets with images containing many objects improves the performance of SR models.

\noindent \textbf{3) Dataset construction.}
To facilitate analysis from the above two perspectives, we introduce a framework that automatically creates a dataset from a set of low-resolution images {collected from the web}.
Specifically, the framework consists of two steps, source selection and object-based filtering, which correspond to the first and second perspectives, respectively. Our framework eliminates the need for manual assessment, facilitating dataset scaling.
In the first step, given several low-resolution image datasets, we estimate the image quality of them through their blockiness distributions and select datasets with quality larger than 90\%.
In the second step, we filter out images with a small number of object regions by using object detection and image segmentation models.
The filtered dataset, which we refer to as the DiverSeg dataset, is utilized for training various SR models in {our} experiments.
We apply our framework to a union set of ImageNet, Places365, and PASS to construct the DiverSeg dataset.
We demonstrate that SR models trained on {DiverSeg} archive state-of-the-art performance.

\section{Related Work}
\label{sec:related_works}

\subsection{Image super-resolution models}

A number of SR models have been proposed that take advantage of deep learning techniques. These models can be divided into two groups, CNN-based models~\cite{dong2014learning, kim2016accurate, dong2016accelerating, shi2016real, ledig2017photo, tong2017image, lim2017enhanced, wang2018esrgan, zhang2018residual, zhang2018image, zhang2019residual} and Transformer-based models~\cite{chen2021pre, li2021efficient, liang2021swinir, zhang2022swinfir, zhang2022efficient, chen2023activating, Zhou_2023_ICCV}. Each group exploits different architectural strengths to enhance low-resolution images to high resolution.


\noindent \textbf{CNN-based models.}
{SRCNN~\cite{dong2014learning} was the first model that integrates a deep convolutional architecture for SR. Subsequently, FSRCNN~\cite{dong2016accelerating} significantly improved computational efficiency by performing convolutional processing in low resolution space and upsampling in the last layer. Furthermore, {ESPCN}~\cite{shi2016real} adopted an efficient {upsampling} method, sub-pixel convolution, to enhance performance while reducing computational costs. These methods established the basic structure of modern SR networks and laid the foundation for subsequent research developments. Later studies introduced various modules such as residual connections~\cite{kim2016accurate,ledig2017photo,lim2017enhanced,wang2018esrgan}, dense blocks~\cite{tong2017image,zhang2018residual,wang2018esrgan}, and attention mechanisms~\cite{zhang2018image, dai2019second, zhang2019residual}. EDSR~\cite{lim2017enhanced} eliminated batch normalization layers and introduced residual scaling to enable stable training of large models. 
MSRResNet~\cite{wang2018esrgan} replaced the basic ResNet blocks with residual dense blocks to improve the balance between performance and computational efficiency. RCAN~\cite{zhang2018image} incorporated channel attention mechanisms to adaptively weight feature representations, significantly enhancing SR performance.}

\noindent \textbf{Transformer-based models.}
{As the first Transformer-based image restoration model, IPT~\cite{chen2021pre} was introduced as a large-scale model utilizing the Transformer's encoder and decoder architecture. By pre-training on ImageNet, IPT significantly improved SR performance, fully leveraging the capabilities of the Transformer. SwinIR~\cite{liang2021swinir} executes self-attention within local windows during feature extraction, demonstrating exceptional SR performance and establishing itself as the foundational model for Transformers in SR. Following this, models that build upon SwinIR have been developed~\cite{zhang2022efficient, zhang2022swinfir, Zhou_2023_ICCV, chen2023activating}, enhancing performance by extending and optimizing the self-attention mechanism. Among them, HAT~\cite{chen2023activating} achieves state-of-the-art performance in SR by using a duplicated cross-attention module and pre-training on ImageNet. However, there may be potential for improvement, as ImageNet is primarily an image recognition dataset and may not be fully optimized for SR.}
\subsection{Image super-resolution datasets}
T91~\cite{4587647} and BSDS200~\cite{martin2001database} are early datasets used to train SR models, consisting of 91 and 200 images, respectively. The turning point in training datasets was ushered by the release of DIV2K~\cite{agustsson2017ntire}, a compilation of 800 high-resolution images with minimal compression noise meticulously collected from the web. Subsequently,  to accommodate the further scaling of the model, Flickr2K~\cite{timofte2017ntire}, consisting of 2,650 high-resolution images, is merged into DIV2K and is referred to as DF2K. In recent years, to further expand the scale of SR datasets, datasets larger than DF2K have been released, such as LSDIR~\cite{Li_2023_CVPR}, comprising 84,991 images and HQ-50K~\cite{yang2023hq} consisting of 50,000 images. These datasets persistently adhere to stringent criteria, ensuring high resolution and negligible compression noise in the imagery. However, since collecting images that meet the aforementioned conditions is quite challenging, these datasets still consist of only tens of thousands of images. In contrast, several approaches~\cite{chen2021pre, li2021efficient, chen2023activating} adopt ImageNet, also known as ImageNet-1k, which consists of 1.28M images spanning diverse categories. These approaches leverage the diversity of texture patterns in ImageNet as an advantage. Nevertheless, ImageNet is pointed out to contain some images having low-resolution and JPEG-compression artifacts, which adversely affect the training results. 

\begin{figure}[t]
\centering
\includegraphics[width=\linewidth]{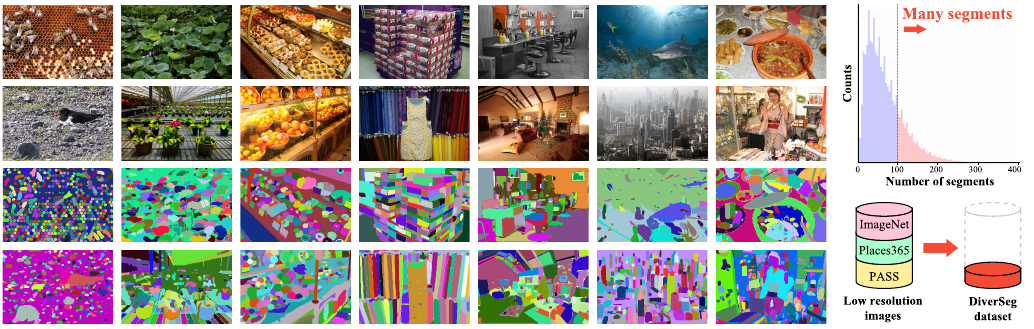}
\figvspacemid
\caption{
Images of DiverSeg with their segmentation masks. DiverSeg is obtained from a large set of low-resolution images through the automated image evaluation pipeline.
}
\label{fig:diverseg}
\figvspacebottom
\end{figure}
\section{DiverSeg dataset}
\label{sec:method}
This section presents  {Diverse Segmentation dataset (DiverSeg)}, our dataset for training SR models without using high-resolution images.
As shown in Figure~\ref{fig:diverseg}, the DiverSeg dataset
consists of low-resolution but high-quality images with diverse segmented object regions.
The framework for constructing the dataset consists of two steps, source selection and object-based filtering, which are designed from our perspective of rethinking resolution and diversity, respectively.
Our approach eliminates the {manual} cost of collecting and quality-checking high-resolution images.

\subsection{Source selection}
\label{sec:selection}

Let $\mathcal{X}$ {be} a set of low-resolution image datasets.
This step filters out low-quality datasets by estimating the quality $\hat{q}_{X} \in [0, 1]$ for each dataset $X \in \mathcal{X}$ and excluding those with $\hat{q}_{X} < 0.9$, under the assumption that low-quality images are detrimental when training SR models.
We introduce a quality estimation method based on the Kullback–Leibler (KL) divergence between blockiness distributions as detailed below.

\noindent \textbf{Image datasets.}
This work uses three web-collected low-resolution datasets $\mathcal{X} = \{\text{ImageNet-1k}, \text{Places365}, \text{PASS}\}$.
Training SR models on them is not straightforward because they may include highly compressed images that negatively affect training of SR models.

\noindent
\textbf{Quality definition.}
Let $Y$ be a dataset of JPEG images. We define the quality $q_{Y}$ by the average JPEG quality, {\it i.e.}, $q_{Y} \triangleq \frac{1}{|Y|}\sum_{y \in Y} Q(y)$ where $0 \leq Q(y) \leq 1$ is the JPEG quality of an image $y$.
The goal of quality estimation is to estimate $q_{X}$ given a dataset $X$. Note that in the estimation phase, datasets may include images other format than JPEG and the true quality $q_{X}$ is not observable.

\begin{figure*}[t]
\centering
\scalebox{1.0}{\includegraphics[width=1\linewidth]{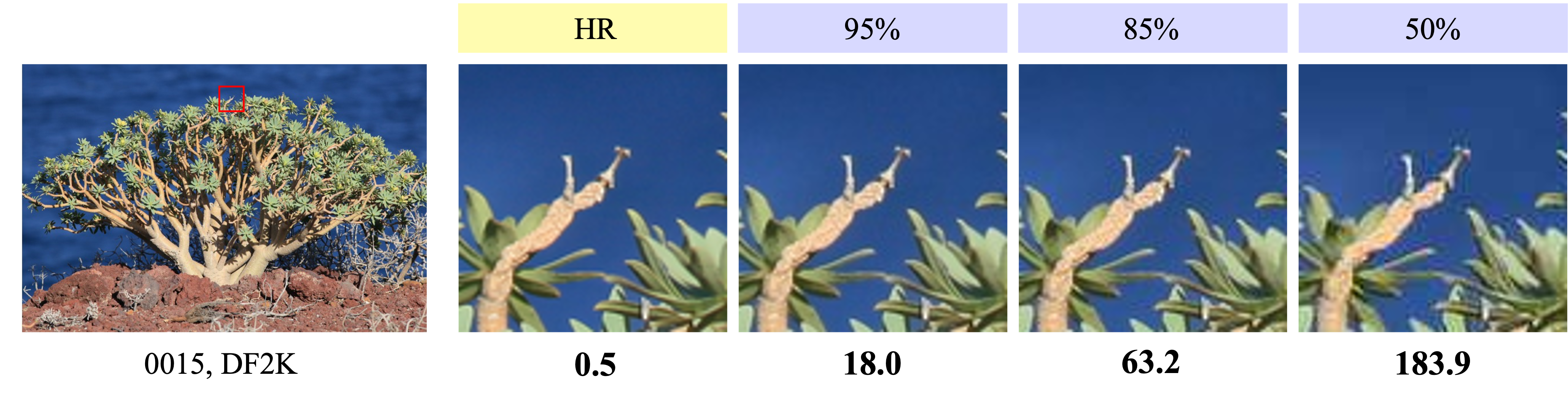}}
\figvspacemid
\vspace{-8pt}
\caption{Comparison of image degradation due to JPEG quality(blue). Blockiness values calculated from the images are marked. As the JPEG quality decreases and artifacts increase, we observe a corresponding rise in blockiness values.}
\label{fig:blockiness}
\figvspacebottom
\end{figure*}
\begin{figure}[t]
\centering
\includegraphics[width=\linewidth]{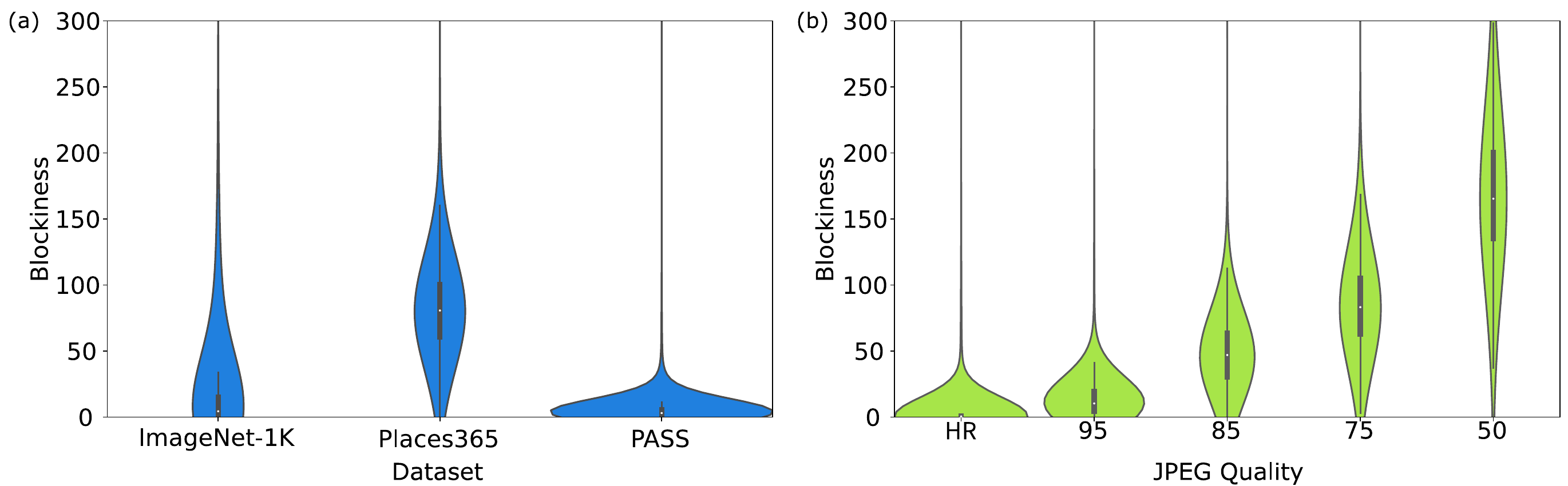}
\figvspacemid
\caption{
(a) Blockiness distributions $p_{X, 1.0}$ for $X = \text{ImageNet-1k}, \text{Places365}$ and $\text{PASS}$.
(b) Basis distributions $p_{Z, q}$ for $Z=\text{DF2K}$ and $q = 0.5, 0.75, 0.85, 0.95, 1.0$.
We estimate the quality by comparing $p_{X,1.0}$ and $p_{Z,q}$ using the KL divergence.}
\figvspacebottom
\label{fig:distribution}
\end{figure}

\noindent
\textbf{Blockiness distribution.}
To estimate the quality, we utilize the blockiness measure~\cite{bhardwaj2018jpeg}. Specifically, for each image dataset $X \in \mathcal{X}$, we estimate the distribution of blockiness values $p_{X,q}(b)$ by kernel density estimation as follows:
\begin{align}
p_{X,q}(b) = \frac{1}{h|X|}\sum_{x \in X} K\left( \frac{b - B(\phi_{q}(x))}{h} \right),
\end{align}
where $x$ is an image, $\phi_{q}$ is the JPEG compression function, $q \in [0, 1]$ is a quality value, $K : \mathbb{R} \to \mathbb{R}$ is a Gaussian kernel, and $h \in \mathbb{R}$ is the bandwidth determined by Scott's method.
The function $B$ is the blockiness measure that measures the quantity of blocking artifacts by computing subband DCT coefficients. Specifically, $B$ is defined on images that are decomposed into $P \times P$ patches as follows:
\begin{align}
B(x) = \sum_{i=1}^{P} \sum_{j=1}^{P} \left| \frac{\bar{V}_{\text{crop}}(i,j) - \bar{V}(i, j)}{\bar{V}(i, j)} \right|,~~
\bar{V}(i,j) = \frac{1}{WH} \sum_{h=1}^{H} \sum_{w=1}^{W} V_{w, h}(i, j)
\end{align}
where $W, H$ are the width and height of an image $x$, $V_{w,h}(i,j)$ is the variation in the $(i, j)$-th subband DCT coefficients within the $(h, w)$-th patch being calculated and its four spatially adjacent patches.
$V_{\text{crop}}(i, j)$ is the variation calculated similarly to $V_{w,h}(i, j)$ for the given image with the first 4 rows and 4 columns removed.
$\bar{V}$ and 
$\bar{V}_{\text{crop}}$ are the average variations of $V_{w,h}(i, j)$ and $V_{\text{crop}}(i, j)$ calculated for each patch, respectively. This work uses $P=8$.
The blockiness value $B(x)$ is expected to be low for uncompressed images and high for compressed images as shown in Figure~\ref{fig:blockiness}.

\noindent
\textbf{Quality estimation.}
We estimate the quality by comparing $p_{X, 1.0}$ with $\{p_{Z, q}\}_{q \in S}$, where $p_{Z, q}$ is a \textit{basis} distribution, a blockiness distribution of images of a fixed quality $q$.
More specifically, the estimated quality is given by
\begin{align}
\hat{q}_{X} = \sum_{q \in S} q \frac{\exp( - D_{\mathrm{KL}}(p_{X, 1.0} || p_{Z, q}))}{\sum_{q^{\prime} \in S} \exp( - D_{\mathrm{KL}}(p_{X, 1.0} || p_{Z, q^{\prime}}))},
\end{align}
where 
$Z$ is a small dataset that involves only uncompressed images. {In this work, we use DF2K.}
$D_\mathrm{KL}$ is the KL divergence, and $S = \{1.0, 0.95, 0.85, 0.75, 0.5\}$ are discretely sampled quality values.
Figure~\ref{fig:distribution} shows the blockiness distributions $p_{X, 1.0}$ for the three low-resolution image datasets and the basis distributions $p_{Z, c}$.

\noindent
\textbf{Selection results.}
The estimated qualities for ImageNet-1k, Places365, and PASS were 
95.5\%, 75.0\% and 99.8\%,
respectively.
From this result, Places365 is filtered out.

\subsection{Object-based filtering}
\label{sec:filtering_methods}

Given a source dataset $X$, this step applies filtering to refine it as a dataset for training SR models, under the assumption that images with diverse object regions are more effective than those with uniform or monotonous content.
Specifically, the refined training dataset is given by $\tilde{X} = \{x \in X: R(x) \geq \theta\}$, where $R(x)$ is the number of object regions and $\theta$ is a threshold.

{We introduce two object-based filtering methods with different granularities to explore how the granularity of object detection, ranging from detailed identification of small objects or features to recognizing larger, more general objects, impacts SR performance.}

\noindent
\textbf{Segmentation-based filtering.}
This method counts the number of object parts by an image segmentation model.
Specifically, we adopt the SAM~\cite{Kirillov_2023_ICCV} with the ViT-H backbone and define $R$ by the number of segmentation masks.
We chose SAM because, unlike typical segmentation models that perform semantic segmentation based on class labels, SAM provides segments that do not impose semantic constraints, allowing for finer region segmentation of diverse objects.
We set $\theta = 100$ as the default, resulting in 260k images remaining after applying the filter to ImageNet-1k.

\noindent
\textbf{Detection-based filtering.}
This method counts the number of objects per image.
We adopt the Detic model~\cite{zhou2022detecting} with the ViT-B backbone and define $R$ by the number of detected objects.
We set $\theta = 18$, resulting in 260k images remaining after applying the filter to ImageNet-1k.


\begin{table}[t]
\centering
\caption{Dataset statistics.
HR: High resolution,
LR: Low resolution,
\#Images: Number of images,
\#Pixels: Average number of pixels per image,
Blockiness: median of blockiness measure indicating the intensity of JPEG compression noise,
\#Segments: Average number of segmentation masks.
}
\resizebox{\linewidth}{!}{
\begin{tabular}{l|c|cc|cccc}
\toprule
Dataset & Task & HR & LR & \#Images  & \#Pixels & Blockiness & \#Segments\\
\midrule
DIV2K~\cite{agustsson2017ntire} & Super-resolution & $\checkmark$ & & 800 & 2.8M & 0.47 & 104 \\
DF2K~\cite{agustsson2017ntire, timofte2017ntire} & Super-resolution & $\checkmark$ & & 3,450      & 2.8M   & 0.47  & 103   \\
LSDIR~\cite{Li_2023_CVPR}
& Super-resolution & $\checkmark$ & & 84,991 & 1.1M & 0.82  & 92  \\
\midrule
ImageNet~\cite{deng2009imagenet} & Image recognition & & $\checkmark$ & {1,281,167}  & 237k     & 4.39  & 71  \\
Places365~\cite{zhou2014learning} & Image recognition & & $\checkmark$ & {1,803,460}  & 366k & 80.71 & 100 \\
PASS~\cite{asano21pass} & Image recognition & & $\checkmark$ & {1,439,589}  & 178k & 3.03  & 74  \\ 
\midrule
DiverSeg-I (Ours) & Super-resolution & & $\checkmark$ & 259,448 & 233k & 2.83 & 146\\
DiverSeg-P (Ours) & Super-resolution & & $\checkmark$ & 267,055 & 179k & 4.39 & 146\\
DiverSeg-IP (Ours) & Super-resolution & & $\checkmark$ & 526,503
& 206k & 3.61 & 146\\
\bottomrule
\end{tabular}
}
\label{tab:data_statistics}
\end{table}

\subsection{Dataset statistics}

Table~\ref{tab:data_statistics} shows dataset statistics.
We created three variants of DiverSeg, namely DiverSeg-I, DiverSeg-P, and DiverSeg-IP, constructed from ImageNet-1k, PASS, and the union of the two, respectively.
The segmentation-based filtering is applied to obtain these datasets. Compared to high-resolution datasets such as DF2K and LSDIR, our datasets have larger number of training images, but contain only low-resolution images.
The median of blockiness values is decreased for ImageNet-1k and is increased for PASS after filtering.

\section{Experiments}

This section conducts experiments by training various SR models on DiverSeg datasets.
We demonstrate that SR models can be trained without using high-resolution images and thoroughly analyze factors from a dataset perspective that are crucial for enhancing the performance of SR.
\begin{table}[t]
\centering
\caption{Comparison of DiverSeg with high-resolution image datasets (DF2K and LSDIR).
Models trained on DiverSeg-I and DiverSeg-P demonstrated superior performance despite not using any high-resolution images for training.}
\label{tab:dataset_comparison}
\resizebox{\linewidth}{!}{
\small
\begin{tabular}{l|l|cc|cc|cc|cc|cc|cc}
\toprule
Model & \multirow{2}{*}{Dataset} & \multirow{2}{*}{HR} & \multirow{2}{*}{LR} & \multicolumn{2}{c|}{Set5} & \multicolumn{2}{c|}{Set14} & \multicolumn{2}{c|}{BSD100} & \multicolumn{2}{c|}{Urban100} & \multicolumn{2}{c}{Manga109} \\
(Params) &&&& PSNR & SSIM & PSNR & SSIM & PSNR & SSIM & PSNR & SSIM & PSNR & SSIM \\
\midrule
                   & DF2K & $\checkmark$ &   & {32.23} & {0.8955}  &  \textbf{{28.67}} & 0.7831   & {27.62} & {0.7374} & 26.23 & 0.7897  & \textbf{30.64} & 0.9108  \\
MSRResNet~\cite{wang2018esrgan}  & LSDIR & $\checkmark$ & & 32.15 & 0.8948  &  {28.66} & {0.7836} & {27.62} & {0.7374}  & \textbf{{26.31}} & \textbf{{0.7918}} & 30.57 & 0.9105 \\
 (1.5M)            & DiverSeg-I && $\checkmark$ & \textbf{32.27} & \textbf{0.8963}  & 28.64  & \textbf{0.7837}   & \textbf{27.64} & \textbf{0.7378}  & \textbf{26.31} & \textbf{0.7918}   & 30.53 & \textbf{0.9115} \\
                   & DiverSeg-P && $\checkmark$ & 32.09 & 0.8943  & 28.61  & 0.7832   & 27.60 & 0.7371  & 26.28 & \textbf{0.7918}   & 30.36 & 0.9101 \\\midrule
                   & DF2K & $\checkmark$ & & 32.50 & 0.8990 & 28.87 & 0.7885  & 27.75 & 0.7421 & 26.73 & 0.8058 & 31.17 & 0.9165 \\
RCAN~\cite{zhang2018image} & LSDIR & $\checkmark$ &   & 32.53 & 0.8992  & 28.89 & 0.7894 & 27.75 & 0.7425 & 26.91 & 0.8090  & 31.33 & 0.9180 \\
(15.5M)            & DiverSeg-I && $\checkmark$ & \textbf{32.70} & \textbf{0.9012}  & \textbf{28.98} & \textbf{0.7908} & \textbf{27.81} & \textbf{0.7443} & \textbf{27.03} & 0.8116  & \textbf{31.58}  & \textbf{0.9210} \\ 
                   & DiverSeg-P && $\checkmark$ & 32.63 & 0.9000  & 28.95 & 0.7898 & 27.77 & 0.7435 & 26.99 & \textbf{0.8134}  & 31.19  & 0.9190 \\\midrule
                   & DF2K & $\checkmark$ &    & {32.61} & 0.8998 & 28.91 & 0.7893  & 27.79 & 0.7434 & 26.84 & 0.8089 & 31.38 & 0.9176 \\
EDSR~\cite{lim2017enhanced}       & LSDIR  & $\checkmark$ &  & 32.57 & 0.8992  & 28.97 & 0.7908  & {27.80} & 0.7438  & {27.05} & {0.8131}  &  {31.47} & 0.9192 \\
(43.0M)            & DiverSeg-I && $\checkmark$ & \textbf{32.71} & \textbf{0.9017}  & 28.98 & 0.7913 & \textbf{27.85} & \textbf{0.7453} & \textbf{27.10} & 0.8142  & \textbf{31.72} & \textbf{0.9216}\\
                   & DiverSeg-P && $\checkmark$ & 32.57 & 0.9002  & \textbf{29.06} & \textbf{0.7915} & 27.80 & 0.7447 & \textbf{27.10} & \textbf{0.8163}  & 31.33 & 0.9191 \\\midrule
                   & DF2K     & $\checkmark$ & & 32.92 & 0.9044 & 29.09 & 0.7950 & 27.92 & 0.7489 & 27.45 & 0.8254 & 32.03 & 0.9260\\
SwinIR~\cite{liang2021swinir}     & LSDIR    & $\checkmark$ & &   32.86 & 0.9036   & 29.16 & {0.7963}   &  27.92 & 0.7492 &   27.79 & 0.8331  &  31.98 & 0.9262 \\
(11.9M)            & DiverSeg-I && $\checkmark$ & \textbf{32.97} & \textbf{0.9053} & 29.23 & \textbf{0.7970} & \textbf{27.98} & \textbf{0.7508} & 27.83 & 0.8336  & \textbf{32.34} & \textbf{0.9283} \\
                   & DiverSeg-P && $\checkmark$ & 32.85 & 0.9040 & \textbf{29.24} & 0.7961 & 27.96 & 0.7502 & \textbf{27.85} & \textbf{0.8349}  & 32.28 & 0.9278 \\\midrule
                   & DF2K     & $\checkmark$ & &   33.03 & 0.9056 & 29.16 & 0.7964 & 27.99 & 0.7514 & 27.93 & 0.8365 & 32.44 & 0.9292 \\
HAT~\cite{chen2023activating}  & LSDIR & $\checkmark$ & & 32.93 & 0.9053 & 29.29 & 0.7988  & 28.01 & 0.7525 & 28.45 & 0.8469 & 32.57 & 0.9306 \\
(20.7M)            & DiverSeg-I & & $\checkmark$ & \textbf{33.15} & \textbf{0.9071} & 29.46 & \textbf{0.8004} & \textbf{28.07} & 0.7542 & 28.51 & 0.8477 & \textbf{32.90} & \textbf{0.9325} \\
                   & DiverSeg-P & & $\checkmark$ & 33.12 & 0.9068 & \textbf{29.50} & 0.8002 & 28.04 & 0.7536 & \textbf{28.53} & \textbf{0.8492} & 32.83 & 0.9320 \\
\bottomrule
\end{tabular}
}
\end{table}
\subsection{Experimental settings}
\label{sec:imp_details}
\noindent
\textbf{SR models.}
We use five models.
Specifically, we use three CNN-based models: MSRResNet~\cite{wang2018esrgan}, EDSR~\cite{lim2017enhanced}, RCAN~\cite{zhang2018image},
and two Transformer-based models: SwinIR~\cite{liang2021swinir} and HAT~\cite{chen2023activating}.

\noindent
\textbf{Training datasets.}
We compare the DiverSeg datasets with two high-resolution datasets: DF2K and LSDIR~\cite{Li_2023_CVPR}.
DF2K is a merged dataset of DIV2K~\cite{agustsson2017ntire} and Flickr2K~\cite{timofte2017ntire}.

\noindent
\textbf{Evaluation datasets.} We use five benchmark datasets: Set5~\cite{bevilacqua2012low}, Set14~\cite{zeyde2012single}, BSD100~\cite{martin2001database}, Urban100~\cite{huang2015single}, and Manga109~\cite{matsui2017sketch}.

\noindent
\textbf{Evaluation metrics.}
PSNR and SSIM on the Y channel (representing luminance) within the transformed YCbCr color space are used as evaluation metrics.

\noindent
\textbf{Implementation settings.}
We follow the training setup of the original papers of the SR models~\cite{wang2018esrgan, lim2017enhanced, zhang2018image, liang2021swinir, chen2023activating}.
Implementation details are provided in the supplementary materials.

\subsection{Experimental results}
\begin{table}[t]
\begin{center}
\caption{Quantitative comparison with state-of-the-art methods on five benchmark datasets. We applied our dataset to two Transformer-based models.
Checkmarks for HR, LR indicate the use of high-resolution and low-resolution datasets, respectively.
}
\label{tab:sota}
\resizebox{\linewidth}{!}{
\begin{tabular}{l|c|cc|cc|cc|cc|cc|cc}
\toprule
\multirow{2}{*}{Method} & \multirow{2}{*}{Training Data} & \multirow{2}{*}{HR} & \multirow{2}{*}{LR} &
\multicolumn{2}{c|}{Set5} &  \multicolumn{2}{c|}{Set14} &  \multicolumn{2}{c|}{BSD100} &  \multicolumn{2}{c|}{Urban100} &  \multicolumn{2}{c}{Manga109}  
\\ 
& & & & PSNR & SSIM & PSNR & SSIM & PSNR & SSIM & PSNR & SSIM & PSNR & SSIM 
\\ 
\midrule
SAN~\cite{dai2019second} & DIV2K & $\checkmark$ & %
& {32.64}
& {0.9003}
& {28.92}
& {0.7888}
& {27.78}
& {0.7436}
& {26.79}
& {0.8068}
& {31.18}
& {0.9169}
\\
IGNN~\cite{zhou2020cross} & DIV2K & $\checkmark$ & %
& {32.57}
& {0.8998}
& {28.85}
& {0.7891}
& {27.77}
& {0.7434}
& {26.84}
& {0.8090}
& {31.28}
& {0.9182}
\\
HAN~\cite{niu2020single} & DIV2K & $\checkmark$ & %
& {32.64}
& {0.9002}
& {28.90}
& {0.7890}
& {27.80}
& {0.7442}
& {26.85}
& {0.8094}
& {31.42}
& {0.9177}
\\
NLSN~\cite{mei2021image} & DIV2K & $\checkmark$ & %
& 32.59 
& 0.9000 
& 28.87 
& 0.7891 
& 27.78 
& 0.7444 
& {26.96}
& {0.8109}
& 31.27 
& 0.9184
\\
RRDB~\cite{wang2018esrgan} & DF2K & $\checkmark$ & %
& {32.73}
& {0.9011}
& {28.99}
& {0.7917}
& {27.85}
& {0.7455}
& {27.03}
& {0.8153}
& {31.66}
& {0.9196}
\\
RCAN-it~\cite{lin2022revisiting} & DF2K & $\checkmark$ & %
& 32.69
& 0.9007
& 28.99
& 0.7922
& 27.87
& 0.7459
& 27.16
& 0.8168
& 31.78
& 0.9217
\\
EDT~\cite{li2021efficient} & DF2K & $\checkmark$ & %
& 32.82
& 0.9031
& 29.09
& 0.7939
& 27.91
& 0.7483
& 27.46
& 0.8246
& 32.05
& 0.9254
\\
HAT-S~\cite{chen2023activating} & DF2K & $\checkmark$ & %
& {32.92}
& {0.9047}
& {29.15}
& {0.7958}
& {27.97}
& {0.7505}
& {27.87}
& {0.8346}
& {32.35}
& {0.9283}
\\
IPT~\cite{chen2021pre} & ImageNet & & $\checkmark$ %
& {32.64}
& {-}
& {29.01}
& {-}
& {27.82}
& {-}
& {27.26}
& {-}
& {-}
& {-}
\\
\midrule
SwinIR~\cite{liang2021swinir} & DF2K & $\checkmark$ & %
& 32.92
& 0.9044
& 29.09
& 0.7950
& 27.92
& 0.7489
& 27.45
& 0.8254
& 32.03
& 0.9260
\\
SwinIR~\cite{liang2021swinir} & DiverSeg-I (Ours) & & $\checkmark$  & \textbf{32.97} & \textbf{0.9053}  & \textbf{29.23} & \textbf{0.7970}  & \textbf{27.98} & \textbf{0.7508} & \textbf{27.83} & \textbf{0.8336} & \textbf{32.34} & \textbf{0.9283} \\
\midrule
HAT~\cite{chen2023activating} & DF2K & $\checkmark$ & %
& {33.04}
& 0.9056
& 29.23
& 0.7973
& 28.00
& 0.7517
& 27.97
& 0.8368
& 32.48
& 0.9292
\\
HAT~\cite{chen2023activating} & ImageNet$\to$DF2K & $\checkmark$ & $\checkmark$ %
& \textbf{33.18}
& \textbf{0.9073}
& 29.38
& 0.8001
& 28.05
& 0.7534
& 28.37
& 0.8447
& 32.87
& 0.9319
\\
HAT~\cite{chen2023activating} & DiverSeg-I (Ours) & & $\checkmark$ %
& 33.15 & 0.9071 & 29.46 & 0.8004 & 28.06 & \textbf{0.7542} & 28.51 & 0.8477 & 32.90 & 0.9325\\
HAT~\cite{chen2023activating} & DiverSeg-IP (Ours) & & $\checkmark$ %
& 33.14 & \textbf{0.9073} & \textbf{29.51} & \textbf{0.8007} & \textbf{28.07} & \textbf{0.7542} & \textbf{28.54} & \textbf{0.8492} & \textbf{32.96} & \textbf{0.9327}\\
\midrule
HAT-L~\cite{chen2023activating} & ImageNet$\to$DF2K & $\checkmark$ & $\checkmark$ %
& \textbf{33.30}
& \textbf{0.9083}
& 29.47
& 0.8015
& 28.09
& 0.7551
& 28.60
& 0.8498
& 33.09
& 0.9335
\\
HAT-L~\cite{chen2023activating} & DiverSeg-I (Ours) & & $\checkmark$ %
& 33.28 & \textbf{0.9083} & 29.54 & \textbf{0.8022} & 28.10 & \textbf{0.7556} & 28.75 & 0.8529  & 33.14 & 0.9340\\
HAT-L~\cite{chen2023activating} & DiverSeg-IP (Ours) & & $\checkmark$ %
& 33.20 & 0.9080 & \textbf{29.59} & 0.8019 &  \textbf{28.11} & \textbf{0.7556} & \textbf{28.81} & \textbf{0.8547} & \textbf{33.19} & \textbf{0.9342}
\\
\bottomrule
\end{tabular}
}
\end{center}
\end{table}

\begin{table}[t]
\centering
\caption{Filtering by blockiness (HAT model, ImageNet-1k). $\theta^{\prime}$ : threshold for blockiness values.}
\label{tab:filters_b}
\tablevspacemid
\scalebox{0.9}{
\small
\begin{tabular}{c|c|cc|cc|cc|cc|cc}
\toprule
\multirow{2}{*}{$\theta^{\prime}$} & \multirow{2}{*}{\#Images} & 
\multicolumn{2}{c|}{Set5} &  \multicolumn{2}{c|}{Set14} &  \multicolumn{2}{c|}{BSD100} &  \multicolumn{2}{c|}{Urban100} &  \multicolumn{2}{c}{Manga109}  
\\ 
& & PSNR & SSIM & PSNR & SSIM & PSNR & SSIM & PSNR & SSIM & PSNR & SSIM 
\\ 
\midrule
10 & 800k & 33.07 & 0.9068 & 29.34 & 0.8000  & 28.04 & \textbf{0.7538} &28.41 & \textbf{0.8465} & 32.56  & 0.9310 \\
30 & 939k  & \textbf{33.13} & \textbf{0.9072} & 29.39 & \textbf{0.8001}  &\textbf{28.06} & \textbf{0.7538}  &\textbf{28.43}  & 0.8463 & 32.70 & 0.9316 \\
100 & 1.08M & 33.11 & 0.9071 & \textbf{29.42} & \textbf{0.8001}  & 28.05 & 0.7537  & 28.41          & 0.8457 & 32.84 & 0.9321 \\
-- & 1.15M & 33.08 & 0.9071 & 29.40 & 0.7999  & 28.05 & 0.7535  & 28.41    & 0.8457 & \textbf{32.88} & \textbf{0.9323} \\
\bottomrule
\end{tabular}}
\end{table}

\noindent
\textbf{Main results.}
To validate the effectiveness of our approach, we trained the five SR models on DiverSeg-I/P.
The results are summarized in Table~\ref{tab:dataset_comparison}.
As shown, models trained on DiverSeg-I/P achieved better performance than those trained on DF2K and LSDIR.
This shows the effectiveness of the proposed datasets.
To the best of our knowledge, we are the first to successfully train SR models without using high-resolution images.

\noindent
\textbf{Comparison with SOTA.}
Table~\ref{tab:sota} shows that DiverSeg improves the state-of-the-art performance. Specifically, the HAT and HAT-L models trained on DiverSeg datasets outperformed those trained with ImageNet-1k$\to$DF2K, which utilizes all ImageNet-1k images for pre-training and the DF2K images for fine-tuning, in terms of PSNR and SSIM on four of five benchmarking datasets.
It is worth noting that DiverSeg-I filters out 77.5\% of images from ImageNet-1k and thus training on it is more efficient than the approach relying on pre-training and fine-tuning.
This confirmed the effectiveness and efficiency of our filtering approach.

\subsection{Analysis 1: Effects of filtering}
\begin{table}[t]
\begin{minipage}{0.54\textwidth}
\centering
\caption{Comparison of filtering methods.
Performance is evaluated using 260k filtered images (HAT model, ImageNet-1k).
}
\tablevspacemid
\vspace{-1pt}
\label{tab:filtering_methods}
\scalebox{0.9}{
\small
\begin{tabular}{l|cc|cc}
\toprule
\multirow{2}{*}{Filtering method} & \multicolumn{2}{c|}{Urban100} & \multicolumn{2}{c}{Manga109} \\
& PSNR & SSIM & PSNR & SSIM \\
\midrule
Blockiness & 28.39 & 0.8467 & 32.47 & 0.9304\\
Detection-based & 28.44 & 0.8462 & 32.87 & 0.9322 \\
Seg.-based & \textbf{28.51} & \textbf{0.8477} & \textbf{32.90} & \textbf{0.9325} \\
\bottomrule
\end{tabular}
}
\end{minipage}
\hfill
\begin{minipage}{0.40\textwidth}
\centering
\caption{Performance comparison across different thresholds $\theta$ (HAT model, ImageNet-1k).}
\label{tab:filtering_threshoulds}
\tablevspacemid
\scalebox{0.78}{
\small
\begin{tabular}{c|c|cc|cc}
\toprule
\multirow{2}{*}{$\theta$} &
\multirow{2}{*}{\#Images} & \multicolumn{2}{c|}{Urban100} & \multicolumn{2}{c}{Manga109} \\
&& PSNR & SSIM & PSNR & SSIM \\
\midrule
0 & 1.2M & 28.41 & 0.8457 & 32.88 & 0.9323 \\
50 & 663k & 28.46 & 0.8472 & \textbf{32.90} & \textbf{0.9325} \\
100 & 259k & \textbf{28.51} & \textbf{0.8477} & \textbf{32.90} & \textbf{0.9325} \\
150 & 86k & 28.36 & 0.8452 & 32.83 & 0.9320 \\
\bottomrule
\end{tabular}
}
\end{minipage}
\end{table}

\begin{table}[t]
\centering
\caption{
Analysis of effects of JPEG quality (HAT model).
}
\tablevspacemid
\label{tab:jpeg_results}
\scalebox{0.78}{
\small
\begin{tabular}{p{1.3cm}|c|c|cc|cc|cc|cc|cc}
\toprule
\multirow{2}{*}{Dataset} & \multirow{2}{*}{Quality (\%)} & \multirow{2}{*}{Blockiness} & \multicolumn{2}{c|}{Set5} & \multicolumn{2}{c|}{Set14} & \multicolumn{2}{c|}{BSD100}& \multicolumn{2}{c|}{Urban100} & \multicolumn{2}{c}{Manga109} \\
&  & & PSNR & SSIM & PSNR & SSIM & PSNR & SSIM & PSNR & SSIM & PSNR & SSIM\\
\midrule
DF2K     & 50    & 165.34 & 31.13 & 0.8855 & 27.54 & 0.7533 & 26.32 & 0.7103 & 25.04 & 0.7511 & 30.20  & 0.9024 \\
         & 75    & 83.20  & 32.63 & 0.9008 & 27.70 & 0.7457 & 27.70 & 0.7457 & 27.39 & 0.8238 & 31.88  & 0.9238 \\
         & 85    & 46.98  & 32.94 & 0.9043 & 29.07 & 0.7941 & 27.93 & 0.7499 & 27.72 & 0.8313 & 32.14 & 0.9266 \\
         & 95    & 10.33  & 32.98 & 0.9048 & 29.11 & 0.7951 & \textbf{27.99} & \textbf{0.7521} & 27.81 & 0.8338 & 32.21 & 0.9278 \\
         & HR    & 0.47   & \textbf{33.03} & \textbf{0.9056} & \textbf{29.16} & \textbf{0.7964} & \textbf{27.99} & 0.7514 & \textbf{27.93} & \textbf{0.8365} & \textbf{32.44} & \textbf{0.9292} \\\midrule
LSDIR    & 50    & 146.43 & 28.50 & 0.8487 & 26.66 & 0.7421 & 24.72 & 0.6674 & 24.91 & 0.7657 & 29.39 & 0.8956  \\
         & 75    & 57.14  & 31.87 & 0.8853 & 28.53 & 0.7805 & 27.49 & 0.7417 & 27.30 & 0.8231 & 31.83 & 0.9223 \\
         & 85    & 24.14  & 32.84 & 0.9044 & 29.17 & 0.7955 & 27.98 & 0.7513 & 28.22 & 0.8412 & 32.38 & 0.9291 \\
         & 95    & 3.71   & \textbf{32.97} & 0.9043 & 29.25 & 0.7979 & \textbf{28.05} & \textbf{0.7539} & 28.37 & 0.8452 & 32.42 & 0.9299 \\
         & HR    & 0.82   & 32.93 & \textbf{0.9053} & \textbf{29.29} & \textbf{0.7988} & 28.01 & 0.7525 & \textbf{28.45} & \textbf{0.8469} & \textbf{32.57} & \textbf{0.9306} \\
\bottomrule
\end{tabular}
}
\end{table}
\noindent
\textbf{Filtering by blockiness measure.}
If the blockiness is an important factor for enchaining SR performance, filtering images by the blockiness values, {\it i.e.}, constructing a training dataset by $\tilde{X} = \{x \in X : B(x) \leq \theta^{\prime}\}$, would be a straightforward approach.
However, as shown in Table~\ref{tab:filters_b},
this filtering did not improve PSNR and SSIM when reducing the threshold $\theta^{\prime}$ from 30 to 10 with an exception of SSIM on Urban100.
This is because this method filters out images without considering the diversity of object regions.

\noindent
\textbf{Object-based filtering.}
Table~\ref{tab:filtering_methods} compares object-based filtering methods described in Sec.~\ref{sec:filtering_methods}, where ImageNet-1k is used as a source dataset.
For a fair comparison, all dataset size after filtering are the same (260k images). 
As shown, the segmentation-based filtering performed the best. This indicates that images with diverse object regions are effective for SR training, and finer granularity leads to better performance.
\begin{figure*}[t]
\centering
\includegraphics[width=0.99\linewidth]{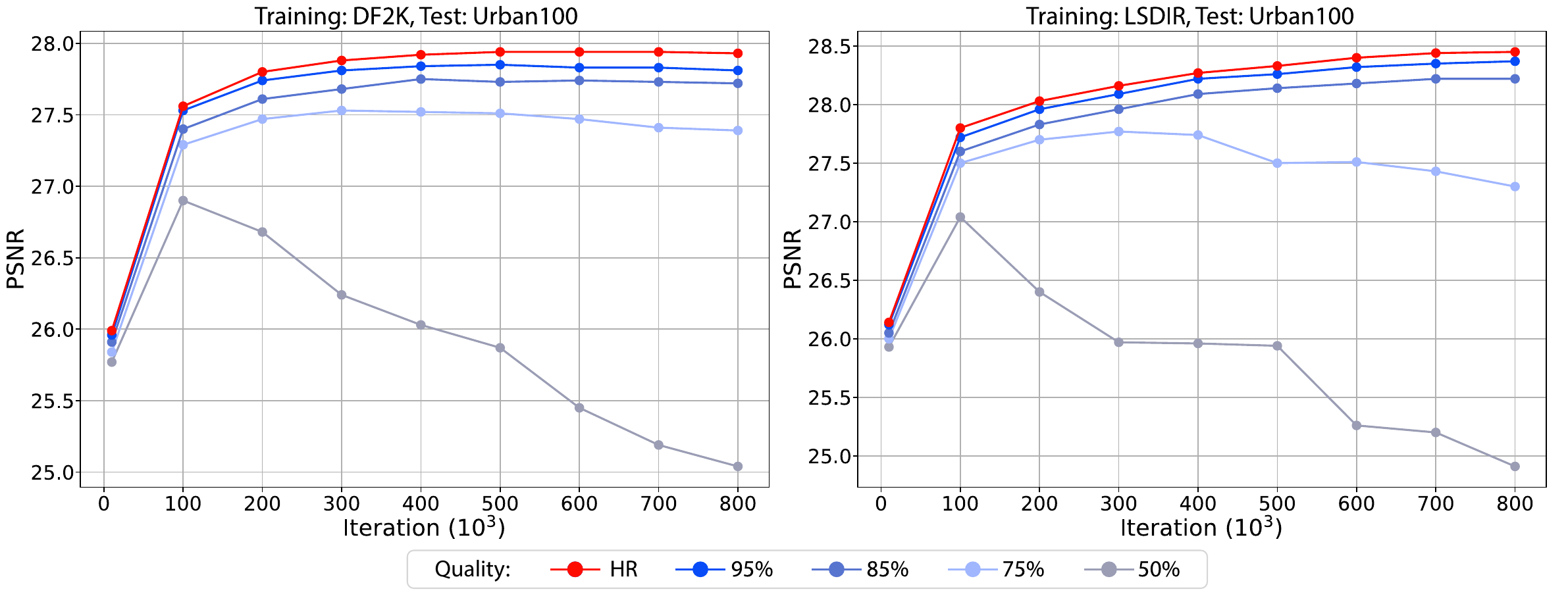}
\figvspacemid
\caption{Comparison of learning processes obtained with various JPEG quality values.
}
\figvspacebottom
\label{fig:jpeg_lr_process}
\end{figure*}
\begin{figure*}[h]
\centering
\includegraphics[width=0.98\linewidth]{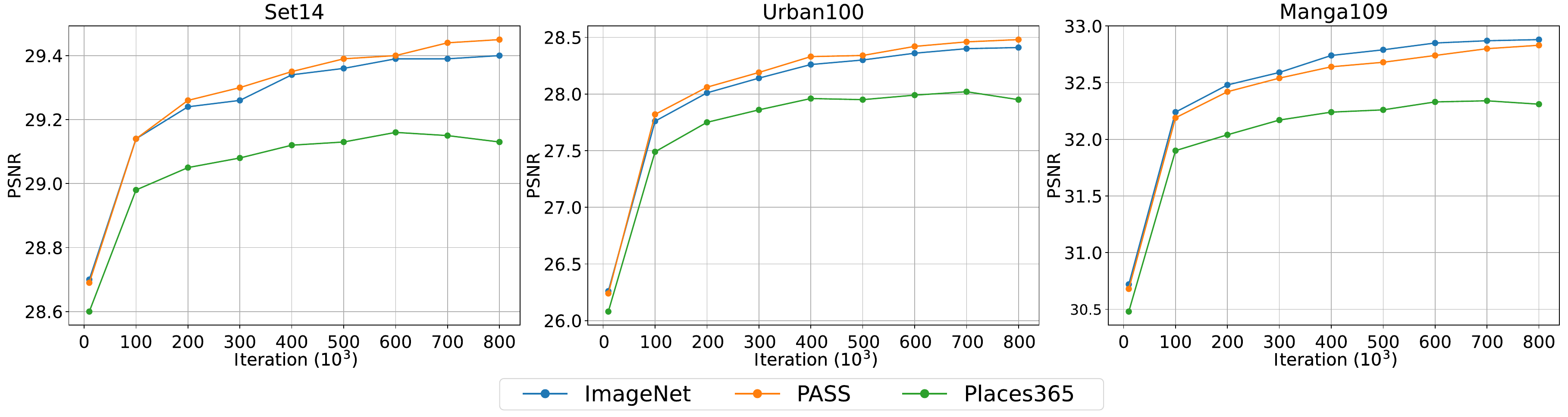}
\figvspacemid
\caption{Comparison of learning processes for ImageNet-1k, PASS, and Places365.
}
\figvspacebottom
\label{fig:dataset_lr_process}
\end{figure*}

\noindent
\textbf{Filtering threshold.}
Table~\ref{tab:filtering_threshoulds} shows the results obtained by varying the threshold $\theta$, which indicates the minimum number of segments.
As shown, $\theta = 100$ performed the best.
\vspace{-10pt}
\subsection{Analysis 2: Effects of image quality}
\label{sec:jpeg_analysis}

In Sec.~\ref{sec:method}, we made an assumption that low-quality images are detrimental when training SR models.
Here, we empirically justify this assumption and evaluate the impact of image quality on SR performance.

\noindent
\textbf{Training with compressed images.}
This experiment applies JPEG compression to the two high-resolution datasets, DF2K and LSDIR, and trains a HAT model on each compressed dataset.
As shown in Table~\ref{tab:jpeg_results}, the performance decreases as the quality decreases on both datasets. 

\noindent
\textbf{Learning process.}
To further analyze why and how low quality images negatively affect training of SR models, Figure~\ref{fig:jpeg_lr_process} compares learning processes.
As can be seen, PSNR decreases after 100k and 300k iterations with 50\% and 75\% qualities, respectively.
With 50\% quality, the final performance at the 800k iteration was worse than that at the 10k iteration.
These results indicate that excluding low-quality images is crucial for enhancing SR performance.
This is in contrast to other vision tasks such as image recognition, where increasing the number of training images, even with low quality images, often helps improve performance.
\begin{figure*}[h]
\centering
\includegraphics[width=\linewidth]{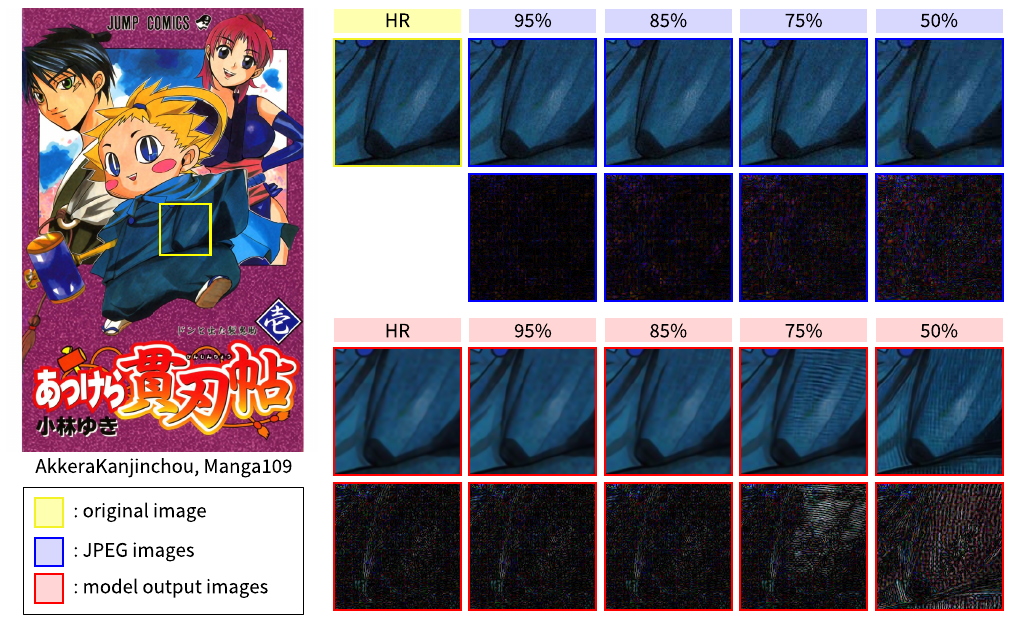}
\figvspacemid
\caption{
Comparison of artifacts produced by JPEG compression (blue) and SR models trained on various JPEG quality values (red, HAT model, DF2K). Numbers indicate JPEG image quality. First and third rows: cropped images. Second and fourth rows: differential images.
}
\figvspacebottom
\label{fig:jpeg_eval}
\end{figure*}
\noindent
\textbf{Training with Places365.}
In the source selection step, the Placses365 dataset was excluded because its quality estimated via the blockiness distribution was low.
To justify this selection, we compare three datasets in Figure~\ref{fig:dataset_lr_process}.
As can be seen, the model trained on Places365 performed worse than those trained on ImageNet-1k and PASS.
Similar to the learning processes obtained from low quality JPEG images, the PSNR decreased in the later phase of the learning process with Places365.
These results confirmed the effectiveness of source selection before training based on the blockiness distributions.
\vspace{-10pt}
\subsection{Analysis 3: Visual comparison}
\begin{figure*}[h]
\centering
\includegraphics[width=\linewidth]{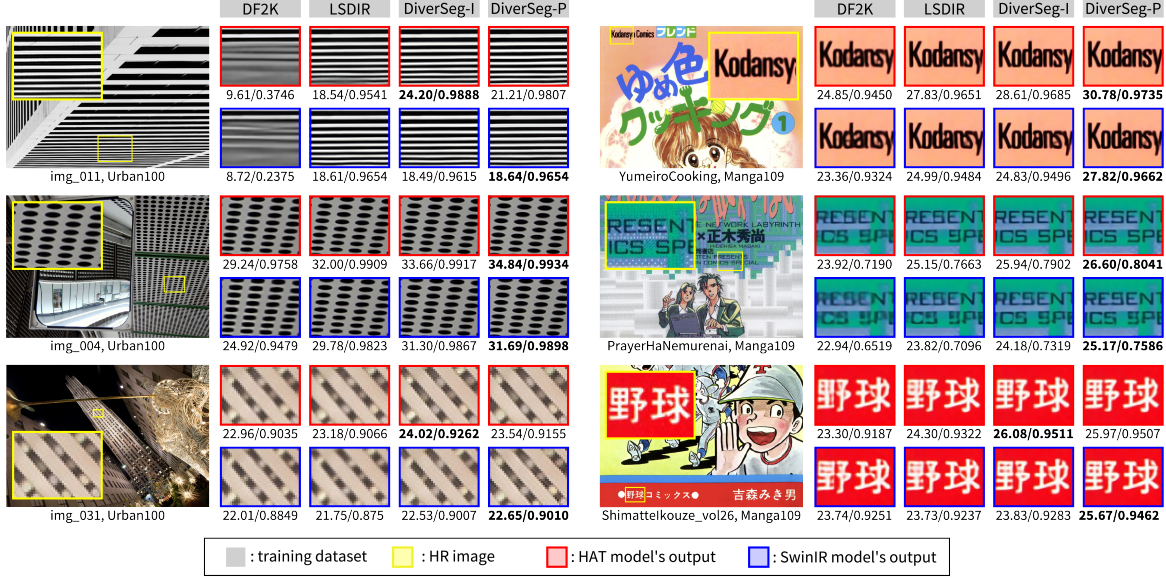}
\figvspacemid
\caption{Visual comparison of $\times 4$ SR models trained on DF2K, LSDIR, DiverSeg-I, and DiverSeg-P datasets. {PSNR/SSIM is calculated for each cropped patch individually to better reflect the differences in performance.}}
\figvspacebottom
\label{fig:main_qualitative}
\end{figure*}
\noindent \textbf{Comparison of artifacts.}
Figure~\ref{fig:jpeg_eval} 
compares the artifacts produced by JPEG compression with those produced by the SR models trained on DF2K using different image quality levels. Images obtained from models trained on 50\% and 75\% quality images show strong stripes or checkerboard patterns of artifacts. These artifacts appear to be more significant than those observed when the original image is compressed using JPEG, suggesting that they are induced by model training. The predisposition to stripes and checkerboard patterns is likely due to the inductive bias inherent in the architecture of the neural network; in particular, the square or rectangular shape of the filters in the convolutional operations could cause this. Improvements to the network architecture to allow training on lower quality images would be interesting as future work.

\noindent \textbf{Qualitative examples.}
Figure~\ref{fig:main_qualitative} shows visual comparisons of SR models trained on DF2K, LSDIR, DiverSeg-I, and DiverSeg-P.
For {the image ``img\_011'' in Urban100}, we observed that models trained on DF2K are unable to recover the horizontal stripes pattern, while the other three models successfully recovered it.
With the three images of Manga109,
we observed that models trained on DiverSeg-I/P exhibit noticeable improvement in the character region compared to those trained on DF2K or LSDIR.
\vspace{-10pt}
\subsection{Discussion and Limitations}
\vspace{-1pt}
\noindent\textbf{High-resolution datasets.}
In this paper, we applied the automated image evaluation pipeline to a large dataset of low-resolution images. We believe our finding contributes to the future construction of training datasets and the development of neural network architectures. Specifically, we demonstrated that SR models can be trained without high-resolution images.
However, this does not imply that high-resolution image datasets are worthless.
Rather, we believe that increasing the diversity of high-resolution images through our proposed filtering method could further improve SR performance.
In Table~\ref{tab:lsdir_filtering}, we examined this on the LSDIR dataset.
Our results show that filtering out images with less than 100 object regions led to increased PSNR and SSIM with the HAT model.
For future work, exploring a hybrid approach that utilizes both low- and high-resolution diverse images could be promising.
\begin{table}[t]
\centering
\caption{Applying object-based filtering to LSDIR with $\theta = 100$ (HAT model).}
\label{tab:lsdir_filtering}
\tablevspacemid
\resizebox{0.95\linewidth}{!}{
\small
\begin{tabular}{l|c|cc|cc|cc|cc|cc}
\toprule
\multirow{2}{*}{Dataset} & \multirow{2}{*}{\#Images} & \multicolumn{2}{c|}{Set5} & \multicolumn{2}{c|}{Set14} & \multicolumn{2}{c|}{BSD100} & \multicolumn{2}{c|}{Urban100} & \multicolumn{2}{c}{Manga109} \\
&& PSNR & SSIM & PSNR & SSIM & PSNR & SSIM & PSNR & SSIM & PSNR & SSIM \\
\midrule
LSDIR & 89,991 & 32.93 & 0.9053 & 29.29 & 0.7988  & 28.01 & 0.7525 & 28.45 & 0.8469 & 32.57 & 0.9306 \\
w/ filtering & 31,561 & \textbf{32.95} & \textbf{0.9056} & \textbf{29.33} & \textbf{0.7991}  & \textbf{28.02} & \textbf{0.7526} & \textbf{28.53} & \textbf{0.8487} & \textbf{32.60} & \textbf{0.9308} \\
\bottomrule
\end{tabular}
}
\end{table}
 


\noindent\textbf{Limitations.} In this study, we focused on two perspectives: resolution and diversity. However, when collecting large datasets, there are other important perspectives, such as fairness and copyright. In addition, for benchmarking purposes, we used the five most commonly used SR datasets for a fair comparison with conventional methods. Nevertheless, real-world applications also face the challenge of blind super-resolution, where the degradation process is unknown. Creating datasets addressing this aspect would be an interesting future research direction.
\vspace{-18pt}
\section{Conclusion}
\vspace{-4pt}
\label{sec:result_discuss}
In this work, we investigated the effect of the training data for SR and showed that SR models are trainable even without using high-resolution images by applying the image evaluation pipeline to a set of large low-resolution images.
In experiments, we thoroughly analyzed the effect of image quality and diversity to SR performance.
We hope that this work will positively influence the future construction of training datasets and lead to better models.

\section*{Acknowledgements}
Computational resource of AI Bridging Cloud Infrastructure (ABCI) provided by National Institute of Advanced Industrial Science and Technology (AIST) was used.

%
%
\bibliographystyle{splncs04}
\bibliography{main}

\begin{thebibliography}{10}
\providecommand{\url}[1]{\texttt{#1}}
\providecommand{\urlprefix}{URL }
\providecommand{\doi}[1]{https://doi.org/#1}

\bibitem{agustsson2017ntire}
Agustsson, E., Timofte, R.: Ntire 2017 challenge on single image super-resolution: Dataset and study. In: CVPRW (2017)

\bibitem{asano21pass}
Asano, Y.M., Rupprecht, C., Zisserman, A., Vedaldi, A.: Pass: An imagenet replacement for self-supervised pretraining without humans. In: NeurIPS Track on Datasets and Benchmarks (2021)

\bibitem{bevilacqua2012low}
Bevilacqua, M., Roumy, A., Guillemot, C., Alberi-Morel, M.L.: Low-complexity single-image super-resolution based on nonnegative neighbor embedding. In: BMVC (2012)

\bibitem{bhardwaj2018jpeg}
Bhardwaj, D., Pankajakshan, V.: A jpeg blocking artifact detector for image forensics. Signal Processing: Image Communication  \textbf{68},  155--161 (2018)

\bibitem{chen2021pre}
Chen, H., Wang, Y., Guo, T., Xu, C., Deng, Y., Liu, Z., Ma, S., Xu, C., Xu, C., Gao, W.: Pre-trained image processing transformer. In: CVPR (2021)

\bibitem{chen2023activating}
Chen, X., Wang, X., Zhou, J., Qiao, Y., Dong, C.: Activating more pixels in image super-resolution transformer. In: CVPR (2023)

\bibitem{dai2019second}
Dai, T., Cai, J., Zhang, Y., Xia, S.T., Zhang, L.: Second-order attention network for single image super-resolution. In: CVPR (2019)

\bibitem{deng2009imagenet}
Deng, J., Dong, W., Socher, R., Li, L.J., Li, K., Fei-Fei, L.: Imagenet: A large-scale hierarchical image database. In: CVPR (2009)

\bibitem{dong2014learning}
Dong, C., Loy, C.C., He, K., Tang, X.: Learning a deep convolutional network for image super-resolution. In: ECCV (2014)

\bibitem{dong2016accelerating}
Dong, C., Loy, C.C., Tang, X.: Accelerating the super-resolution convolutional neural network. In: ECCV (2016)

\bibitem{huang2015single}
Huang, J.B., Singh, A., Ahuja, N.: Single image super-resolution from transformed self-exemplars. In: CVPR (2015)

\bibitem{kim2016accurate}
Kim, J., Lee, J.K., Lee, K.M.: Accurate image super-resolution using very deep convolutional networks. In: CVPR (2016)

\bibitem{Kirillov_2023_ICCV}
Kirillov, A., Mintun, E., Ravi, N., Mao, H., Rolland, C., Gustafson, L., Xiao, T., Whitehead, S., Berg, A.C., Lo, W.Y., Dollar, P., Girshick, R.: Segment anything. In: ICCV (2023)

\bibitem{ledig2017photo}
Ledig, C., Theis, L., Husz{\'a}r, F., Caballero, J., Cunningham, A., Acosta, A., Aitken, A., Tejani, A., Totz, J., Wang, Z., et~al.: Photo-realistic single image super-resolution using a generative adversarial network. In: CVPR (2017)

\bibitem{li2023feature}
Li, A., Zhang, L., Liu, Y., Zhu, C.: Feature modulation transformer: Cross-refinement of global representation via high-frequency prior for image super-resolution. In: CVPR (2023)

\bibitem{li2021efficient}
Li, W., Lu, X., Qian, S., Lu, J., Zhang, X., Jia, J.: On efficient transformer-based image pre-training for low-level vision. arXiv preprint arXiv:2112.10175  (2021)

\bibitem{Li_2023_CVPR}
Li, Y., Zhang, K., Liang, J., Cao, J., Liu, C., Gong, R., Zhang, Y., Tang, H., Liu, Y., Demandolx, D., Ranjan, R., Timofte, R., Van~Gool, L.: Lsdir: A large scale dataset for image restoration. In: CVPRW (2023)

\bibitem{li2023ntire}
Li, Y., Zhang, Y., Timofte, R., Van~Gool, L., Yu, L., Li, Y., Li, X., Jiang, T., Wu, Q., Han, M., et~al.: Ntire 2023 challenge on efficient super-resolution: Methods and results. In: CVPRW (2023)

\bibitem{liang2021swinir}
Liang, J., Cao, J., Sun, G., Zhang, K., Van~Gool, L., Timofte, R.: Swinir: Image restoration using swin transformer. In: ICCVW (2021)

\bibitem{lim2017enhanced}
Lim, B., Son, S., Kim, H., Nah, S., Mu~Lee, K.: Enhanced deep residual networks for single image super-resolution. In: CVPRW (2017)

\bibitem{lin2022revisiting}
Lin, Z., Garg, P., Banerjee, A., Magid, S.A., Sun, D., Zhang, Y., Van~Gool, L., Wei, D., Pfister, H.: Revisiting rcan: Improved training for image super-resolution. arXiv preprint arXiv:2201.11279  (2022)

\bibitem{martin2001database}
Martin, D., Fowlkes, C., Tal, D., Malik, J.: A database of human segmented natural images and its application to evaluating segmentation algorithms and measuring ecological statistics. In: ICCV (2001)

\bibitem{matsui2017sketch}
Matsui, Y., Ito, K., Aramaki, Y., Fujimoto, A., Ogawa, T., Yamasaki, T., Aizawa, K.: Sketch-based manga retrieval using manga109 dataset. Multimedia Tools and Applications  \textbf{76},  21811--21838 (2017)

\bibitem{mei2021image}
Mei, Y., Fan, Y., Zhou, Y.: Image super-resolution with non-local sparse attention. In: CVPR (2021)

\bibitem{niu2020single}
Niu, B., Wen, W., Ren, W., Zhang, X., Yang, L., Wang, S., Zhang, K., Cao, X., Shen, H.: Single image super-resolution via a holistic attention network. In: ECCV (2020)

\bibitem{shi2016real}
Shi, W., Caballero, J., Husz{\'a}r, F., Totz, J., Aitken, A.P., Bishop, R., Rueckert, D., Wang, Z.: Real-time single image and video super-resolution using an efficient sub-pixel convolutional neural network. In: CVPR (2016)

\bibitem{timofte2017ntire}
Timofte, R., Agustsson, E., Van~Gool, L., Yang, M.H., Zhang, L.: Ntire 2017 challenge on single image super-resolution: Methods and results. In: CVPRW (2017)

\bibitem{tong2017image}
Tong, T., Li, G., Liu, X., Gao, Q.: Image super-resolution using dense skip connections. In: ICCV (2017)

\bibitem{wang2018esrgan}
Wang, X., Yu, K., Wu, S., Gu, J., Liu, Y., Dong, C., Qiao, Y., Change~Loy, C.: Esrgan: Enhanced super-resolution generative adversarial networks. In: ECCVW (2018)

\bibitem{4587647}
Yang, J., Wright, J., Huang, T., Ma, Y.: Image super-resolution as sparse representation of raw image patches. In: CVPR (2008)

\bibitem{yang2023hq}
Yang, Q., Chen, D., Tan, Z., Liu, Q., Chu, Q., Bao, J., Yuan, L., Hua, G., Yu, N.: Hq-50k: A large-scale, high-quality dataset for image restoration. arXiv preprint arXiv:2306.05390  (2023)

\bibitem{zeyde2012single}
Zeyde, R., Elad, M., Protter, M.: On single image scale-up using sparse-representations. In: Proc. 7th Int. Conf. Curves Surf. (2010)

\bibitem{zhang2022swinfir}
Zhang, D., Huang, F., Liu, S., Wang, X., Jin, Z.: Swinfir: Revisiting the swinir with fast fourier convolution and improved training for image super-resolution. arXiv preprint arXiv:2208.11247  (2022)

\bibitem{zhang2022efficient}
Zhang, X., Zeng, H., Guo, S., Zhang, L.: Efficient long-range attention network for image super-resolution. In: ECCV (2022)

\bibitem{zhang2019residual}
Zhang, Y., Li, K., Zhong, B., Fu, Y.: Residual non-local attention networks for image restoration. In: ICLR (2019)

\bibitem{zhang2018image}
Zhang, Y., Li, K., Li, K., Wang, L., Zhong, B., Fu, Y.: Image super-resolution using very deep residual channel attention networks. In: ECCV (2018)

\bibitem{zhang2018residual}
Zhang, Y., Tian, Y., Kong, Y., Zhong, B., Fu, Y.: Residual dense network for image super-resolution. In: CVPR (2018)

\bibitem{zhang2023ntire}
Zhang, Y., Zhang, K., Chen, Z., Li, Y., Timofte, R., Zhang, J., Zhang, K., Peng, R., Ma, Y., Jia, L., et~al.: Ntire 2023 challenge on image super-resolution (x4): Methods and results. In: CVPRW (2023)

\bibitem{zhou2014learning}
Zhou, B., Lapedriza, A., Xiao, J., Torralba, A., Oliva, A.: Learning deep features for scene recognition using places database. Advances in neural information processing systems  \textbf{27} (2014)

\bibitem{zhou2020cross}
Zhou, S., Zhang, J., Zuo, W., Loy, C.C.: Cross-scale internal graph neural network for image super-resolution. Advances in neural information processing systems  \textbf{33},  3499--3509 (2020)

\bibitem{zhou2022detecting}
Zhou, X., Girdhar, R., Joulin, A., Kr{\"a}henb{\"u}hl, P., Misra, I.: Detecting twenty-thousand classes using image-level supervision. In: ECCV (2022)

\bibitem{Zhou_2023_ICCV}
Zhou, Y., Li, Z., Guo, C.L., Bai, S., Cheng, M.M., Hou, Q.: Srformer: Permuted self-attention for single image super-resolution. In: ICCV (2023)

\end{thebibliography}

\clearpage

\end{document}